\documentclass{article}
\usepackage[margin=2.5cm]{geometry}
\usepackage{authblk}

\usepackage{microtype}
\usepackage{graphicx}
\usepackage{subfigure}
\usepackage{booktabs} 

\usepackage{hyperref}

\usepackage{amsmath}
\usepackage{amssymb}
\usepackage{mathtools}
\usepackage{amsthm}

\usepackage{algorithm}
\usepackage{algpseudocode}

\usepackage[capitalize,noabbrev]{cleveref}

\theoremstyle{plain}

\theoremstyle{definition}

\theoremstyle{remark}

\usepackage[textsize=tiny]{todonotes}

\usepackage[commandnameprefix=always]{changes}

\definechangesauthor[name={Shirin Dora}, color=red]{SD}
\definechangesauthor[name={Dylan Adams}, color=green]{DA}

\newcommand{\algoAbbr}{IMSNN}
\newcommand{\algoAbbrPlural}{IMSNNs}

\title{Synaptic Modulation using Interspike Intervals Increases Energy Efficiency of Spiking Neural Networks}
\author[1]{Dylan Adams}
\author[1]{Magda Zajaczkowska}
\author[2]{Ashiq Anjum}
\author[1]{Andrea Soltoggio}
\author[1]{\authorcr Shirin Dora \thanks{Corresponding author: s.dora@lboro.ac.uk}}
\affil[1]{\small Department of Computer Science, Loughborough University, Loughborough, United Kingdom}
\affil[2]{School of Computing and Mathematic Sciences, University of Leicester, Leicester, United Kingdom}
\date{}

\begin{document}

\maketitle











\begin{abstract}
Despite basic differences between Spiking Neural Networks (SNN) and Artificial Neural Networks (ANN), most research on SNNs involve adapting ANN-based methods for SNNs. Pruning (dropping connections) and quantization (reducing precision) are often used to improve energy efficiency of SNNs. These methods are very effective for ANNs whose energy needs are determined by signals transmitted on synapses. However, the event-driven paradigm in SNNs implies that energy is consumed by spikes. In this paper, we propose a new synapse model whose weights are modulated by Interspike Intervals (ISI) i.e. time difference between two spikes. SNNs composed of this synapse model, termed ISI Modulated SNNs (\algoAbbr), can use gradient descent to estimate how the ISI of a neuron changes after updating its synaptic parameters. A higher ISI implies fewer spikes and vice-versa. The learning algorithm for \algoAbbrPlural{} exploits this information to selectively propagate gradients such that learning is achieved by increasing the ISIs resulting in a network that generates fewer spikes. The performance of \algoAbbrPlural{} with dense and convolutional layers have been evaluated in terms of classification accuracy and the number of spikes using the MNIST and FashionMNIST datasets. The performance comparison with conventional SNNs shows that \algoAbbrPlural{} exhibit upto 90\% reduction in the number of spikes while maintaining similar classification accuracy.

\end{abstract}

\section{Introduction}

Artificial Neural Networks (ANN) have progressed greatly in recent years, enabling the development of highly effective solutions for many problems such as image recognition \cite{Krizhevsky2012} and natural language processing \cite{Wolf2020}. ANNs are generally deployed on energy-intensive devices with graphical processing units to speed-up their computations. The growing popularity of ANNs has led to greater interest in development of solutions that can operate in energy-constrained environments including mobile phones and drones.

\begin{figure}[t]
    \centering
    \includegraphics[width=8cm]{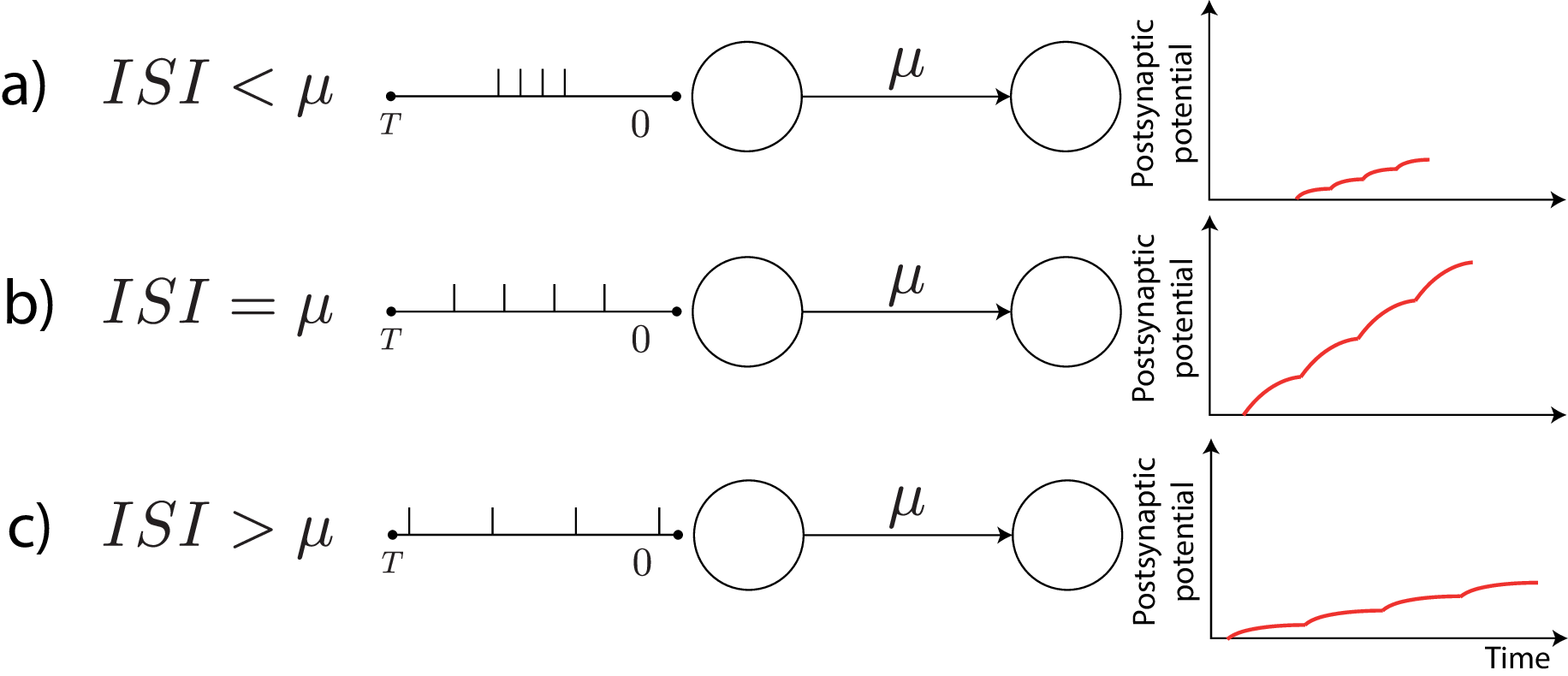}
    \label{fig:synapse-idea}
    \caption{In \algoAbbrPlural{}, the contribution of a spike to the postsynaptic potential depends on the Interspike Interval (ISI) of the presynaptic neurons. a), b) and c) show the postsynaptic potential for three different scenarios, namely \(ISI < \mu\), \(ISI = \mu\) and \(ISI > \mu\), respectively. The presynaptic pattern in each case has 4 spikes but the postsynaptic potential contributed by these spikes is different across the three scenarios. The synaptic parameter \( \mu \) represents the ISI that result in maximum postsynaptic potential (see b)). As ISI deviates from \( \mu \), the contribution to postsynaptic potential goes down (see a) and c)).}
\end{figure}

In recent years, Spiking Neural Networks (SNN) have emerged as an energy-efficient alternative to ANNs. Inspired by biological neurons, SNNs are composed of spiking neurons which communicate using binary events called spikes at specific time instants. A spiking neuron integrates any incoming spikes into its membrane potential and transmit a spike to downstream neurons whenever this membrane potential surpasses a defined threshold value \cite{Gerstner2014}. This event driven computational paradigm underlies low power processing in the brain and enables energy-efficient computing \cite{Krichmar2019}.

Current research in SNNs is strongly motivated by adapting the techniques developed for ANNs to build SNNs. Particularly, the effectiveness of gradient-based approaches for ANNs has inspired several error-backpropagation methods for SNNs \cite{Bohte2002,Lee2016,Shrestha2018,Machingal2023}. The key issue is that the gradient of a spike with respect to a neuron's potential doesn't exist. Surrogate gradient methods overcome this issue by using a well-behaved surrogate function to replace this derivative \cite{Neftci2019}. These methods have helped bridge the gap between the performance of ANNs and SNNs, and intensified research into further improving the energy efficiency of SNNs.

Current approaches to energy-efficiency for SNNs can be divided in two categories based on the goal of optimization, namely \textit{compression} and \textit{spike cardinality} methods. Compression methods reduce the size of the network through techniques like pruning \cite{Neftci2016, Rathi2017, Wu2019b, Martinelli2020, Lew2023} and quantization \cite{Chowdhury2021, Yan2021, Li2022}. Spike cardinality methods focus on reducing the number of spikes generated within the network. Pruning and quantization methods for SNNs have been adapted from ANNs where number of connections and the precision of weights determine the energy requirements. However, the event-driven paradigm in SNNs implies that energy is consumed only when a spike is generated. Thus, spike cardinality methods are arguably the natural and more effective way of reducing energy required by SNNs.

AutoSNN \cite{Na2022} uses evolutionary methods to search for SNNs that generate fewer spikes and exhibit high performance. However, training and evaluating a large number of networks is computationally expensive approach to build energy efficient SNNs. In this paper, we propose an algorithmic method to develop SNNs that generate fewer spikes.



We have developed a new synapse model whose weight is a function of the Interspike Intervals (ISI) of the presynaptic neuron (Figure \ref{fig:synapse-idea}). The parameters of the synapse represent the presynaptic ISI that will result in maximum contribution to the postsynaptic neuron. As the presynaptic ISI deviates from the synaptic parameter, spikes result in lower contribution to the postsynaptic neuron. The SNNs composed of these synapses, termed ISI Modulated SNNs (\algoAbbr), use gradient descent to estimate how a parameter update affects the ISI of the postsynaptic neuron. The learning algorithm for \algoAbbrPlural{} exploits this information to propagate gradients selectively. Specifically, it uses an adapted learning rule so that learning doesn't result in lowering of ISIs in the nentwork. Preventing a lowering of ISI restricts an increase in the number of spikes generated in the network. 

The performance of \algoAbbrPlural{} have been evaluated in terms of classification accuracy and the number of spikes generated within the network using MNIST and FashionMNIST dataset. The results of performance evaluation have been compared with that of conventional SNNs which uses synapses with a fixed weight for the whole simulation. The classification accuracy of \algoAbbrPlural{} is similar to that of conventional SNNs while the number of spikes generated by \algoAbbrPlural{} is upto 90\% less than those generated in conventional SNNs.

Rest of the paper is organized as follows. Section \ref{sec:methods} presents the new synapse model and the learning algorithm for \algoAbbrPlural{}. \ref{sec:results} presents the results of performance evaluation for \algoAbbrPlural{}.


\section{Related Works}
\label{sec:related-worsks}
\noindent \textbf{Compression Methods}: Compression methods include techniques that optimize connectivity using methods like pruning and quantizationto to reduce the size of the network.

Pruning methods focus on reducing the size of the network by dropping those connections whose removal from the network does not affect its performance significantly. Differnt criteria have been proposed for pruning connections which include thresholding \cite{Neftci2016,Rathi2017, Liu2020a}, difference between output spike trains \cite{Wu2019b}, importance of connections for performance \cite{Martinelli2020} and algorithmically learning connectivity \cite{Chen2021}.

Quantization methods reduce the precision of learned network parameters thereby lowering the number of bits in arithmatic operation. In \cite{Rathi2017}, a 2-layer SNN is trained using unsupervised Spike Timing Dependent Plasticity (STDP). After training, connections with weights below a threshold value are pruned. The remaining weights in the network are set to their average value to obtain a network with 2-level quantization.

\textbf{Spike cardinality methods}: Spike cardinality methods directly target reducing the number of spikes generated in the network. Inspired by neural architecture search \cite{Real2017}, AutoSNN \cite{Na2022} is an evolutionary framework to seach for SNN architectures with high accuracy and low spike counts. AutoSNN employs a primary network with placeholder blocks that can be substituted with spiking convolutional layers and spiking residual layers. The search algorithm generates and evaluates 100 architectures derived from the primary network.

A related direction of research is to reduce the number of time steps required by SNNs \cite{Mueller2021, Bu2023} for simulation which may not necessarily reduce the number of spikes in the network.


\section{Methods}
\label{sec:methods}
In this section, we first describe forward propagation in \algoAbbrPlural{} and the proposed synapse model. Second, we present the error-backpropagation based learning algorithm for \algoAbbrPlural{} which reduces the number of spikes generated by the trained networks.

\subsection{Forward Propagation in \algoAbbr}
\label{sec:forward-prop}
Consider a SNN with $ L \in \mathbb{N}^+ $ fully connected layers where layer $ l \in  \{ 1, \cdots, L \}$ consists of $ \mathcal{N}^{(l)} $ spiking neurons. All neurons except output layer neurons are modeled using the Leaky Integrate and Fire neuron model \cite{PerezNieves2021}. $ v^{(l)}_j(t) \in \mathbb{R}$ denotes the membrane potential of the neuron \(j\) in layer \(l\) at time step \( t \). The neuron emits a spike \( s^{(l)}_j (t) \)  at time \(t\) when its membrane potential crosses a threshold \(\theta\) given as
\begin{equation}
    s^{(l)}_j(t) = \begin{cases}
        1 & \text{if } v^{(l)}_j(t) \geq \theta \\
        0 & \text{otherwise}  \\
    \end{cases}
\end{equation}

$\theta$ is set to 1 for all neurons in the network, except output layer neurons. The output layer neurons do not generate a spike which implies that they continuously accumulate their potential during the simulation. The membrane potential of a neuron evolves in
time according to the following equation
\begin{equation}
    v^{(l)}_j(t+1) = \beta v^{(l)}_j(t) + \sum_i s^{(l-1)}_i(t) \vartheta^{(l-1)}_{ij}(t)
    \label{eq:neuron_rolled}
\end{equation}
where $\beta$ is the membrane decay constant. $ \vartheta^{(l-1)}_{ij}(t) $ is the weight of the connection between neuron $i$ in layer $(l-1)$ and neuron $j$ in layer $l$ at time \(t\). The weight of a connection at time \(t\) is estimated based on the time elapsed since the last presynaptic spike (see Section \ref{sec:synapse-model}). The membrane potential is reset to $ 0 $ when a neuron emits a spike. 

The predicted class for a given input spike pattern is determined based on the membrane potential accumulated by the output neurons at time $T$. The probability $(p_j)$ that a sample belongs to the class associated with the output neuron $j$ is given as
\begin{equation}
    p_j = \frac{v^L_j (T)}{\sum^{n^{(L)}}_{h = 1} v^L_h (T)}
\end{equation}


\subsection{Synapse Model}
\label{sec:synapse-model}

\begin{figure}
    \centering
    \includegraphics[width=8cm]{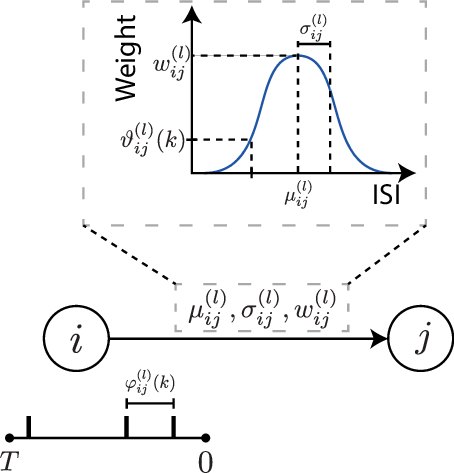}
    \caption{Proposed synapse model used in \algoAbbrPlural. \( \mu^{(l)}_{ij}, \sigma^{(l)}_{ij} \) and \( w^{(l)}_{ij} \) represent the mean, width and height of the synapse, respectively. Note that only heights are learnt in this paper. Mean and width are assigned random values at the start of training and are not learnt during training.}
    \label{fig:synapse_model}
\end{figure}

Synapses in \algoAbbrPlural{} regulate the propagation of presynaptic spikes by modulating the potential contributed by these spikes to the postsynaptic neurons. The extent of modulation for each presynaptic spike depends on the parameters of synapses and the Spike Frequency (SF) of presynaptic neurons. The parameters of synapses can be learned to propagate presynaptic spike patters with specific SFs.

Figure \ref{fig:synapse_model} shows a schematic illustration of the synapse model. \algoAbbrPlural{} utilize the Inter-spike Interval (ISI) to parameterize the SFs transmitted on a synapse. ISI is the time difference betwee two consecutive spikes. At the start of a simulation, the ISI of all neurons is initialized to 0. The ISI \( (\varphi^{(l)}_{i}) \) of a neuron at time \(t\) is given by
\begin{equation}
    \varphi^{(l)}_i (t+1) = 1 + \varphi^{(l)}_i (t) (1 - s^{(l)}_i (t))
\end{equation}

We model the relationship between the weight of a connection at time \(t\) and the ISI using a Gaussian function, given by
\begin{equation}
    \vartheta^{(l)}_{ij} (t) = w^{(l)}_{ij} \exp\left( {-\frac{\left(\varphi^{(l)}_{i} (t) - \mu^{(l)}_{ij} \right)^2}{2 \left( \sigma^{(l)}_{ij} \right)^2}}\right)
    \label{eq:synapse_model}
\end{equation}
where \( w^{(l)}_{ij} \), \( \mu^{(l)}_{ij} \) and \( \sigma^{(l)}_{ij} \) are height, mean and width, respectively of the Gaussian function associated with the synapse between neuron \(i\) in layer \((l-1)\) and neuron \(j\) in layer \(l\). For brevity, \( w^{(l)}_{ij} \), \( \mu^{(l)}_{ij} \) and \( \sigma^{(l)}_{ij} \) will be referred as height, mean and width of a synapse, respectively in the rest of the paper. The impact of \( w^{(l)}_{ij} \) on the postsynaptic potential is similar to a synaptic weight in conventional SNNs (see \ref{sec:demonstration}). 

It can be observed from Equation \eqref{eq:synapse_model} that the weight of a connection is highest when the presynaptic spikes have an ISI of \( \mu^{(l)}_{ij} \). In this case, a spike results in maximum potential being contributed to the postsynaptic neuron. The weight of the synapse becomes smaller as the difference between the ISI and \( \mu^{(l)}_{ij} \) increases, resulting in lower contribution to the postsynaptic membrane potential. This allows the synapses in \algoAbbrPlural{} to propagate spike patterns with specific SFs.

For all results reported in this paper, mean and width of synapses are initialized in the interval $[5, 10]$ and $[10, 50]$ respectively. Further, mean and width are not learnt during training. Heights are always initialized in the interval $[-0.05, 0.05]$ and learned using the algorithm presented in Section \ref{sec:learning-algorithm}.

\subsection{Learning Algorithm}
\label{sec:learning-algorithm}
In this subsection, we first derive the Error-Backpropagation (EBP) based learning rule for updating heights of synapses in \algoAbbrPlural{}. We then highlight that the learning rule contains a gradient term which represents how updating a particular height affects the ISI of the postsynaptic neuron. This information is utilised to adapt the EBP-based learning rule for those synapses whose update reduces the ISI of the postsynaptic neuron. A lower ISI implies more spikes are generated by neurons in the network.

As in \cite{PerezNieves2021}, we apply backpropagation through time using the unrolled form of a neuron's potential in time. The unrolled potential of a neuron can be obtained from \eqref{eq:neuron_rolled} and is given as
\begin{equation}
    v^{(l)}_j(t+1) = \sum^{\mathcal{N}^{(l-1)}}_{i=1} \left( \sum_{k=0}^{t}  \beta^{t-k}  s^{(l-1)}_i(k) \vartheta^{(l-1)}_{ij}(k)  \right)
    \label{eq:neuron_model}
\end{equation}

Applying backpropagation through time, the gradient for updating the height of a synapse is given as
\begin{align}
    \nabla w^{(l)}_{ij} = \sum_t &\underbrace{\epsilon^{(l+1)}_j (t)}_{\substack{\text{Gradient from} \\ \text{next layer}}}  \overbrace{\frac{d s^{(l+1)}_j (t)}{d v^{(l+1)}_j (t)}}^{\substack{\text{Spike} \\ \text{derivative}}}  \nonumber  \\
    &\underbrace{\left( \sum_{k<t}  \beta^{t-1-k}  s^{(l)}_i (k)  \frac{ d \vartheta^{(l)}_{ij} (k) }{ d w^{(l)}_{ij} }  \right)}_{\text{Input trace}}  \label{eq:update-rule-without-adjustment}
\end{align}
The update rule in \eqref{eq:update-rule-without-adjustment} can be understood as a product of three terms over all time steps. The \textit{gradient from next layer} \( (\epsilon^{(l+1)}_j) \) is the derivative of the loss \((\mathcal{L})\) with respect to the spike output of the postsynaptic neuron \(j\). The second term is the derivative of the postsynaptic spikes with respect to the potential of the postsynaptic neuron at time \(t\). As this derivative is not defined, we use a surrogate gradient function proposed in this paper \cite{Zenke2018, Zenke2021}. The third term is the derivative of the postsynaptic potential with respect to the height of the synapse. The derivative in the third term can be obtained from \eqref{eq:synapse_model} (see Appendix \ref{sec:d-weight-by-d-height}).

For output layer neurons, the gradient  from the next layer \( (\epsilon^{(L)}_j) \) in \eqref{eq:update-rule-without-adjustment} is the derivative of the loss with respect to the response of output neurons. For other layers, \( \epsilon^{(l+1)}_j (t) \) can be defined recursively (see Appendix \ref{sec:recursive-definition-epsilon} for a derivation) as
\begin{equation}
    \epsilon^{(l+1)}_j (t) =  \sum^{\mathcal{N}^{(l+2)}}_{h=1}  \sum_{k>t}  \epsilon^{(l+2)}_h (k)  \overbrace{\frac{ d s^{(l+2)}_h (k) }{ d v^{(l+2)}_h (k) }}^{\substack{\text{Spike} \\ \text{derivative}}}  \underbrace{\frac{ dv^{(l+2)}_h (k) }{ ds^{(l+1)}_j (t) }}_{\substack{\text{Potential} \\ \text{derivative}}}  \label{eq:next-layer-gradient}
\end{equation}

For conventional SNNs, the \textit{potential derivative} in \eqref{eq:next-layer-gradient} will be a function of the weight of the synapse and \( \beta \) (see Appendix \eqref{sec:epsilon-conventional-SNN}). In \algoAbbrPlural{}, the potential derivative also depends on ISI because weight is a function of ISI (see \eqref{eq:synapse_model}). Thus, we can compute the gradient of loss with respect to ISI \( ( \nabla \varphi^{(l+1)}_{j} (t) ) \) and estimate how updating a particular height will alter the ISI of the postsynaptic neuron. The equation for computing \( \nabla \varphi^{(l+1)}_j (t) \) is given in Appendix \ref{sec:ISI-gradient}. A negative value for \( \nabla \varphi^{(l+1)}_{j} (t) \) indicates that updating the heights will also increase the postsynaptic ISI resulting in fewer spikes and vice-versa. 

The learning algorithm for \algoAbbrPlural{} exploits this information to propagate gradients selectively by adapting the potential derivative in \eqref{eq:next-layer-gradient}. Synapses with negative \( \nabla \varphi^{(l+1)}_{j} (t) \) are updated using the actual potential derivative as they increase the ISI of the postsynaptic neuron. For synapses with positive \( \nabla \varphi^{(l+1)}_{j} (t) \), the componet of potential derivative containing the gradient with respect to \( \varphi^{(l+1)}_{j} (t) \) is suppressed. The actual potential derivative in \eqref{eq:next-layer-gradient} is given by (see Appendix \ref{sec:potential-derivative} for derivation)
\begin{align}
    \frac{ dv^{(l+2)}_h (k) }{ ds^{(l+1)}_j (t) } &= 
    \beta^{k-1-t}  \vartheta_{jh} (t)  +  \nonumber  \\
    &\sum\limits^{k-1}_{m=t+1}  \beta^{k-1-m}  s_j (m)  \frac{d \vartheta_{jh} (m)}{d \varphi_j (m)}  \frac{d \varphi_{jh} (m)}{d s_j (t)}
    \label{eq:potential-derivative}
\end{align}
where the derivative with respect to \( \varphi^j (m) \) is adapted as
\begin{equation}
    \frac{d \vartheta^{(l+1)}_{jh} (m)}{d \varphi^{(l+1)}_j (m)} = \begin{cases}
        - \frac{\left(  \varphi_{jh} (m) - \mu_{jh}  \right)  \vartheta_{jh} (m)}{\sigma_{jh}^2}  &  \nabla \varphi_{j} (m) < 0 \\
        0  &  \nabla \varphi_{j} (m) \geq 0
    \end{cases}
    \label{eq:rho}
\end{equation}
For brevity, the superscript \( (l+1) \) has been dropped on the right side of \eqref{eq:potential-derivative} and \eqref{eq:rho}. It may be noted from \eqref{eq:potential-derivative} and \eqref{eq:rho} that the updates for neurons with positive \(\nabla \varphi_{j} (m)\) are not completely suppressed. This allows the gradients to flow from deeper to earlier layers of the network.


Equations \eqref{eq:update-rule-without-adjustment}, \eqref{eq:next-layer-gradient}, \eqref{eq:potential-derivative} and \eqref{eq:rho} together represent the learning rule for \algoAbbrPlural{}. Algorithm \ref{alg:learning-algorithm} shows the pseudocode for the proposed learning algorithm.

\begin{algorithm}[t]
    \caption{Learning Algorithm for \algoAbbrPlural}
    \label{alg:learning-algorithm}
 \begin{algorithmic}
    \State {\bfseries Input:} \algoAbbr{} with \( L \) layers; \( E \) epochs; initialize \( \mu \), \( \sigma \) and \( w \) for all synapses
    \For{\( e = 1 \) {\bfseries to} \( e = E \)}
    \For{\( l = 1 \) {\bfseries to} \( l = (L - 1) \)}
    \For {\( j, h \)}
    \If{\( \nabla \varphi^{(l+1)}_j (m) \geq 0\)}
    \State \( \frac{d \vartheta^{(l+1)}_{jh} (m)}{d \varphi^{(l+1)}_j (m)} = 0\)
    \EndIf
    \EndFor
    \State Update height \( w^{(l)}_{ij} \)
    \EndFor
    \EndFor
 \end{algorithmic}
 \end{algorithm}

\section{Results} 
\label{sec:results}
In this section, first, we demonstrate the working of the proposed synapse model using a single LIF neuron. Second, we present the results of an ablation study to understand the impact of proposed learning algorithm on the number of spikes generated in the network. Third, to verify the effectiveness of our method, the performance of \algoAbbrPlural{} is evaluated on two benchmark datasets using fully connected networks and spiking convolution networks. 

The performance of \algoAbbrPlural{} is compared with the performance of conventional SNNs with same architecture and initialization settings. The decay constant $(\beta)$ of the neuron in all models is set to $ 0.99 $. The weights and heights of SNNs and \algoAbbrPlural{}, respectively are randomly initialized using the normal distribution with mean 0 and standard deviation of 0.05. 

Cross entropy loss $(\mathcal{L})$ is used to train all models reported in this paper. The performance of the models presented in this section has been evaluated using two metrics, namely classification accuracy and average number of spikes generated in a given layer of the network, respectively. The metric for classification accuracy \( (\kappa_a) \) is given as
\begin{equation}
    \kappa_a = \frac{\text{\# Correctly classified samples}}{\text{\# Samples}} \times 100
\end{equation}

As in previous works \cite{Kundu2021}, the average spike count \( (\kappa_n^{(l)}) \) of a layer $l$ is the is computed as the ratio of the total spike count in $T$ steps over all the neurons of the layer $l$ to the total number of neurons in that layer
\begin{equation}
    \kappa_n^{(l)} = \frac{\sum_t \sum_i s^{(l)}_i (t)}{\mathcal{N}^{(l)}}
\end{equation}
A single metric for the whole network is obtained by summing \( \kappa_n^{(l)} \) for all layers in the network.

The results for fully connected SNNs and \algoAbbrPlural{} presented in this section are based on five repeated runs using a single architecture. For spiking convolutional networks, the results are based on single run using the given architecture. All models are trained using the Adam Optimizer for 20 epochs using a learning rate of 1e-4. Each sample is presented to the network for a duration of \( T = 100\)ms with a time step of 1ms. Samples from all datasets are normalized to the interval \( [0, 1] \) and are encoded into spikes using firing rates in the range \( [28.5, 100] \) Hertz. Each sample is presented to the network for a duration of 100ms.

\begin{figure}[t]
    \centering
    \includegraphics[scale=0.4]{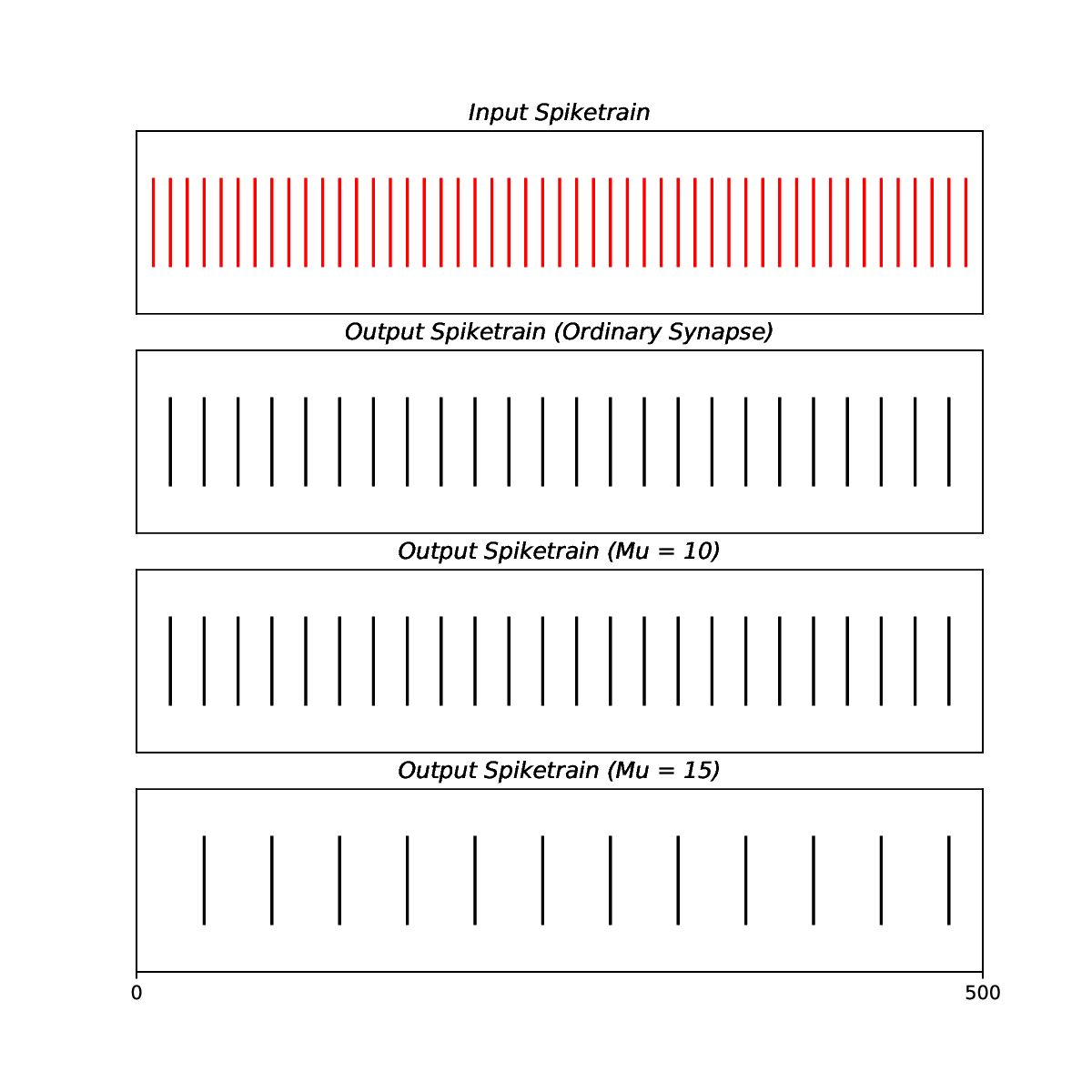}
    \caption{Output of LIF neurons with a single input synapse in three networks. First network has a conventional synapse with fixed weight throughout the simulation. Second and third networks use the proposed synapse with \( \mu \) set to 10ms and 15 ms, respectively.}
    \label{fig:single-LIF-noiseless}
\end{figure}

\begin{table*}[t]
    \centering
    \caption{Performance comparison of \algoAbbr vs SNN in networks with one and two hidden layers on MNIST Dataset}
    \begin{tabular}{|l|l|c|c|c|}
        \hline
        Model & Architecture &  Train & Test & \# Spikes per neuron \\
         &  &  Accuracy (\%) & Accuracy (\%) & in each layer \( (\kappa^{(l)}_n) \)\\
        \hline
        \hline
        SNN & 784-500-10 & 99.54 (0.02) & 97.43 (0.38)& 1.19 (0.04)\\
        \algoAbbr & 784-500-10 & 99.73 (0.03) & 97.45 (0.02) & 0.17 (0.01)\\
        \hline
        SNN & 784-500-500-10 & 99.26 (0.07) & 97.66 (0.18) & 0.90 (0.04), 1.03 (0.08) \\
        \algoAbbr & 784-500-500-10 & 99.31 (0.10) & 97.40 (0.20) & 0.07 (0.00), 0.34 (0.02) \\
        \hline
    \end{tabular}
    \label{tab:mnist-results}
\end{table*}

\begin{table}[b]
    \centering
    \caption{Comparison of networks obtained by suppressing negative and positive gradients of the loss with respect to ISI on MNIST datasets}
    \begin{tabular}{|l|l|c|c|}
        \hline
        Model & Architecture & \# Spikes per \\
         &  & neuron \( (\kappa^{(l)}_n) \) \\
        \hline
        \hline
        \algoAbbr{} & 784-500-10 & 0.17 \\
        SNN & 784-500-10 & 1.19 \\ 
        \algoAbbr{}\( ^c \) & 784-500-10 & 36.88 \\
        \hline
    \end{tabular}
    \label{table:ablation}
\end{table}

\subsection{Demonstration of the Synapse Model}
\label{sec:demonstration}
This section illustrates the functioning of the proposed synapse model using networks with single LIF neurons having a single input synapse. Three separate networks are simulated in this section. The first network has a conventional synapse whose weight is set to 0.6 throughout the simulation. Second and third networks use the proposed synapse model with $\mu$ set to 10ms and 15ms, respectively. $\sigma$ is set to 5 in both the second and third networks. The height \( (w) \) of the gaussian function in second and third networks is set to 0.6, which is equal to the weight in the first network. The weight of the synapses in the second and third networks are estimated using Equation \ref{eq:synapse_model}.

Figure \ref{fig:single-LIF-noiseless} presents the output spike trains generated by the LIF neurons in the three networks when an input spike pattern with an ISI of 10ms is presented via their input synapse. It can be observed that the output of the network with \( \mu = 10ms \) is exactly same as the response of the SNN with a conventional synapse. This is because \( \mu \) is equal to ISI, as a result, the weight of the synapse in the second network is always equal to the height of the gaussian function which is equal to weight of synapse in the first network.

Further, the number of output spikes in the second and third networks is different even though \( w \) is set to the same value. There are more output spikes generated in the third network due to smaller difference between presynaptic ISI and mean ISI of the synapse. This exhibits the capabilities of the proposed synapse model to propagate presynaptic spikes with specific spike frequencies.

\subsection{Ablation Study}
The learning algorithm in Section \ref{sec:learning-algorithm} set the positive gradients of loss with respect to ISI to zero for reducing spikes generated in the network. A complimentry hypothesis is that setting negative gradients to zero would increase the number of spikes generated in the network. In this section, we conducted an ablation study to validate this hypothesis by setting the negative gradients to zero during training on the MNIST dataset.

Table \ref{table:ablation} shows the results of comparison between a conventional SNN and networks trained using the learning algorithm presented in \ref{sec:learning-algorithm} and the complimentary hypothesis above (denoted by \( \text{\algoAbbr{}}^c \)). The accuracy of the three networks is similar but there is a significant difference in the mean number of spikes generated by neurons. The network trained using the complimentary hypothesis resulted in significantly more spikes than the other two networks.

\begin{table*}[t]
    \centering
    \caption{Performance comparison of \algoAbbr vs SNN in networks with one and two hidden layers on FashionMNIST Dataset}
    \begin{tabular}{|l|l|c|c|c|}
        \hline
        Model & Architecture &  Train & Test & \# Spikes per neuron \\
         &  &  Accuracy (\%) & Accuracy (\%) & in each layer \( (\kappa^{(l)}_n) \)\\
        \hline
        \hline
        SNN & 784-500-10 & 98.30 (0.06)& 86.11 (0.30)& 0.49 (0.03) \\
        \algoAbbr & 784-500-10 & 91.55 (0.17) & 88.27 (0.11) & 0.22 (0.00) \\
        \hline
        SNN & 784-500-500-10 & 97.32 (0.09) & 86.65 (0.14) & 1.53 (0.17), 0.51 (0.04)\\
        \algoAbbr & 784-500-500-10 & 92.90 (0.17) & 86.01 (0.09) & 0.07 (0.00), 0.30 (0.02)\\
        \hline
        \multicolumn{5}{|c|}{Spiking Convolutional Networks} \\
        \hline
        SNN & 784-48c5-8c5-500-10 & 95.2 & 88.05 & 0.4, 1.81, 0.54 \\
        \algoAbbr & 784-48c5-8c5-500-10 & 89.11 & 85.47 & 0.04, 0.15, 0.37 \\
        \hline
    \end{tabular}
    \label{tab:fashionmnist-results}
\end{table*}

\subsection{Performance Evaluation}
In this section, the results of performance evaluation of conventional SNNs and \algoAbbrPlural{} on the MNIST and FashionMNIST datasets are presented. The numbers in parentheses in the Tables \ref{tab:mnist-results} and \ref{tab:fashionmnist-results} represent the standard deviations for corresponding results.

\textbf{MNIST}: Table \ref{tab:mnist-results} presents the results of performance evaluation using networks with one and two hidden layers. The last column in the table shows the number of spikes generated by neurons in each layer (separated by `,') of the network except the input and output layers. The activity of neurons in the input layer is determined by encoding and output layer neurons do not generate spikes (see section \ref{sec:forward-prop}). It can be observed that all the networks have similar classification accuracy during training and testing. But, both \algoAbbrPlural{} generate significantly fewer spikes compared to conventional SNNs. There is a reduction of 86\% and 79\% (approximately) in the number of spikes generated in the networks with one and two hidden layers respectively. For the network with two hidden layers, there is a reduction of 92\% and 67\% in the two hidden layers, respectively. The lower reduction in deeper layer can be attributed to the fact that the activity in deeper layers depend on the activity in earlier layers of the network.

\textbf{FashionMNIST}: Table \ref{tab:fashionmnist-results} shows the results of performance evaluation for \algoAbbrPlural{} and conventional SNNs on the FashionMNIST dataset. All networks in the table exhibit similar classification accuracies with \algoAbbrPlural{} using significantly fewer spikes to achieve the reported performance. Compared to SNNs, \algoAbbrPlural{} with one and two hidden layers generated 55\% and 92\% fewer spikes, respectively. There is a reduction of 95\% and 41\% in the number of spikes generated in the first and second hidden layer, respectively. For the FashionMNIST dataset, the table also shows results for two networks with convolutional layers. For reducing the number of model parameters, the mean and width in convolutional \algoAbbrPlural{} at all locations within a single kernel have been assigned identical values. It can be observed from the table that neurons in \algoAbbr{} generated 80\% fewer spikes compared to conventional SNNs while achieving similar classification accuracy.

\section{Conclusion}
This paper develops a new SNN, termed ISI Modulated SNN (\algoAbbr) that can be optimized to use fewer spikes during training. \algoAbbrPlural{} employ a new synapse model whose weight depends on the ISIs of presynaptic spikes. The functional relationship between weights and ISIs makes it possible to estimate how updating the parameters of a synapse affect the ISI of the postsynaptic neurons. The learning algorithm for \algoAbbrPlural{} uses the gradient of loss with respect to ISI to selectively update the synaptic parameters such that the ISIs of the postsynaptic neurons have high ISIs after training. Higher ISIs result in a SNN that generates fewer spikes. The performance of \algoAbbrPlural{} is evaluated using classification accuracy and the number of spikes using MNIST and FashionMNIST datasets. The results clearly indicate that \algoAbbrPlural{} achieve similar accuracy to conventional SNNs while generating upto 90\% lesser spikes. Future work on \algoAbbrPlural{} will focus on developing a better reasoning for variability in the number of spikes generated in networks with different architectures.

\bibliography{imsnn}
\bibliographystyle{unsrt}

\newpage
\appendix
\onecolumn

\section{Deriving Recursive Definition of \( \epsilon^{(l+1)}_j \)}
\label{sec:recursive-definition-epsilon}
Based on EBP, the update rule for heights in \algoAbbrPlural{} is given by
\begin{align}
    \nabla w^{(l)}_{ij} = \sum_t \underbrace{\epsilon^{(l+1)}_j (t)}_{\substack{\text{Gradient from} \\ \text{next layer}}}  \overbrace{\frac{d s^{(l+1)}_j (t)}{d v^{(l+1)}_j (t)}}^{\substack{\text{Spike} \\ \text{derivative}}} \underbrace{\left( \sum_{k<t}  \beta^{t-1-k}  s^{(l)}_i (k)  \frac{ d \vartheta^{(l)}_{ij} (k) }{ d w^{(l)}_{ij} }  \right)}_{\text{Input trace}}  \label{eq:update-rule-without-adjustment-appendix}
\end{align}
By definition, \( \epsilon^{(l+1)}_j \) is given by
\begin{equation}
    \epsilon^{(l+1)}_j (t) = \frac{ d \mathcal{L} }{ d s^{(l+1)}_j (t) }
    \label{eq:epsilon-appendix}
\end{equation}
Applying chain rule in reverse, i.e. starting from \(s^{(l+1)}_j\), to the above equation
\begin{align}
    \epsilon^{(l+1)}_j (t)  &=  \sum^{\mathcal{N}^{(l+2)}}_{h=1}  \sum_{k>t}  \frac{ d \mathcal{L} }{ d v^{(l+2)}_h (k) }  \frac{ d v^{(l+2)}_h (k) }{ d s^{(l+1)}_j (t) }  \nonumber  \\
    &=  \sum^{\mathcal{N}^{(l+2)}}_{h=1}  \sum_{k>t}  \frac{ d \mathcal{L} }{ d s^{(l+2)}_h (k) }  \frac{ d s^{(l+2)}_h (k) }{ d v^{(l+2)}_h (k) }  \frac{ d v^{(l+2)}_h (k) }{ d s^{(l+1)}_j (t) }  \nonumber  \\
    &=  \sum^{\mathcal{N}^{(l+2)}}_{h=1}  \sum_{k>t}  \epsilon^{(l+2)}_h (k)  \frac{ d s^{(l+2)}_h (k) }{ d v^{(l+2)}_h (k) }  \frac{ d v^{(l+2)}_h (k) }{ d s^{(l+1)}_j (t) }
    \label{eq:epsilon-appendix}
\end{align}
In the above equation, the summation over time is for \( k > t \) because a spike generated by the neuron \( j \) at time \( t \) will affect the potential of neurons in layer \( (l + 2) \) at all time steps in the future.

\section{\( \epsilon^{(l+1)}_j \) for Conventional SNNs}
\label{sec:epsilon-conventional-SNN}
In conventional SNNs, the weight of a connection is fixed during the simulation. Therefore, the unrolled form of the potential is given by 
\begin{equation}
    v^{(l)}_j(t+1) = \sum^{\mathcal{N}^{(l-1)}}_{i=1} \left( \sum_{k=0}^{t}  \beta^{t-k}  s^{(l-1)}_i(k) \omega^{(l-1)}_{ij}  \right)
    \label{eq:neuron-model-SNN}
\end{equation} 
where \( \omega^{(l-1)}_{ij}\) denotes the weight of a connection. Therefore, the derivative of the potential at time \( k \) with respect to a spike at time \( t \) (last derivative in \eqref{eq:epsilon-appendix}) is given by
\begin{equation}
    \frac{ d v^{(l+2)}_h (k) }{ d s^{(l+1)}_j (t) } = \beta^{k-1-t} \omega^{(l-1)}_{jh}
\end{equation}
Thus, \eqref{eq:epsilon-appendix} for conventional SNNs reduces to
\begin{align}
    \epsilon^{(l+1)}_j (t) =  \sum^{\mathcal{N}^{(l+2)}}_{h=1}  \sum_{k>t}  \epsilon^{(l+2)}_h  \frac{ d s^{(l+2)}_h (k) }{ d v^{(l+2)}_h (k) }  \beta^{k-1-t} \omega^{(l-1)}_{jh}
    \label{eq:epsilon-SNN-appendix}
\end{align}

\section{Derivative of Weight in \algoAbbrPlural}
\label{sec:d-weight-by-d-height}
The derivative of the weight of a synapse in \algoAbbrPlural{} with respect to height of the gaussian is given as
\begin{equation}
    \frac{ d \vartheta^{(l)}_{ij} (k) }{ d w^{(l)}_{ij} } = \exp  \left(  - \frac{ \left( \varphi^{(l)}_{i} (k) - \mu^{(l)}_{ij} \right)^2 }{ 2 \left( \sigma^{(l)}_{ij} \right)^2}  \right)  \label{eq:d-weight-by-d-height}
\end{equation}

\section{Potential Derivative in \algoAbbrPlural}
\label{sec:potential-derivative}
Consider the scenario where we are updating the height \( w^{(l)}_{ij} \). The potential derivative in \eqref{eq:next-layer-gradient} is given by

\begin{align}
    \frac{d v^{(l+2)}_h (k) }{ d s^{(l+1)}_j (t)} &= \sum^{k-1}_{m=t} \frac{d \left( \beta^{k-1-m} s^{(l+1)}_j (m) \vartheta^{(l+1)}_{jh} (m) \right)}{d s^{(l+1)}_j (t)}  \label{eq:potential-derivative-1}  \\
    &= \beta^{k-1-t}  \vartheta^{(l+1)}_{jh} (t)  +  \sum^{k-1}_{m=t+1}  \beta^{k-1-m}  s^{(l+1)}_j (m)  \frac{d \vartheta^{(l+1)}_{jh} (m)}{d s^{(l+1)}_j (t)}  \label{eq:potential-derivative-2}  \\
    &= \beta^{k-1-t}  \vartheta^{(l+1)}_{jh} (t)  +  \sum^{k-1}_{m=t+1}  \beta^{k-1-m}  s^{(l+1)}_j (m)  \frac{d \vartheta^{(l+1)}_{jh} (m)}{d \varphi^{(l+1)}_j (m)}  \frac{d \varphi^{(l+1)}_{jh} (m)}{d s^{(l+1)}_j (t)}  \label{eq:potential-derivative-3}
\end{align}




\section{ISI Gradient \( (\nabla \varphi^{(l+1)}_j (t)) \)}
\label{sec:ISI-gradient}
\( \nabla \varphi^{(l+1)}_j (t) \) can also be expressed in terms of gradients from next layer. Applying the chain rule in reverse, \( \nabla \varphi^{(l+1)}_j (t) \) is given by
\begin{align}
    \nabla \varphi^{(l+1)}_j (t) &= \sum^{\mathcal{N}^{(l+2)}}_{h=1}  \frac{d \mathcal{L}}{d \vartheta^{(l+1)}_{jh} (t)}  \frac{d \vartheta^{(l+1)}_{jh} (t)}{d \varphi^{(l+1)}_j (t)}  \nonumber  \\
    &= \sum^{\mathcal{N}^{(l+2)}}_{h=1}  \sum_{k>t}  \frac{d \mathcal{L}}{d v^{(l+2)}_{h} (k)}  \frac{d v^{(l+2)}_{h} (k)}{d \vartheta^{(l+1)}_{jh} (t)}  \frac{d \vartheta^{(l+1)}_{jh} (t)}{d \varphi^{(l+1)}_j (t)}  \nonumber  \\
    &= \sum^{\mathcal{N}^{(l+2)}}_{h=1}  \sum_{k>t}  \frac{ d \mathcal{L} }{ d s^{(l+2)}_h (k) }  \frac{ d s^{(l+2)}_h (k) }{ d v^{(l+2)}_h (k) }  \frac{d v^{(l+2)}_{h} (k)}{d \vartheta^{(l+1)}_{jh} (t)}  \frac{d \vartheta^{(l+1)}_{jh} (t)}{d \varphi^{(l+1)}_j (t)}  \nonumber  \\
    &= \sum^{\mathcal{N}^{(l+2)}}_{h=1}  \sum_{k>t}  \epsilon^{(l+2)}_h (k)  \frac{ d s^{(l+2)}_h (k) }{ d v^{(l+2)}_h (k) }  \frac{d v^{(l+2)}_{h} (k)}{d \vartheta^{(l+1)}_{jh} (t)}  \frac{d \vartheta^{(l+1)}_{jh} (t)}{d \varphi^{(l+1)}_j (t)}
\end{align}

\end{document}